\documentclass[11pt]{article}

\usepackage[preprint]{acl}

\usepackage{times}
\usepackage{latexsym}

\usepackage[T1]{fontenc}

\usepackage[utf8]{inputenc}

\usepackage{microtype}

\usepackage{inconsolata}

\usepackage{graphicx}
\usepackage{tcolorbox}
\tcbuselibrary{breakable}
\usepackage{booktabs}
\usepackage{amsmath}
\usepackage{multirow} 
\usepackage{amsfonts}
\usepackage{makecell}
\newcommand{\bi}[1]{\textbf{\textit{#1}}}
\title{Pragmatic Inference for Improving Social Reasoning Generalization in\\Large Language Models}

\newcommand{\IU}{\textsuperscript{\(\spadesuit\)}}
\newcommand{\NTU}{\textsuperscript{\(\heartsuit\)}}
\newcommand{\UM}{\textsuperscript{\(\diamondsuit\)}}
\newcommand{\QC}{\textsuperscript{\(\clubsuit\)}}
\newcommand{\MSU}{\textsuperscript{\(\triangle\)}}
\newcommand{\UCSB}{\textsuperscript{\(\star\)}}

\author{
  \textbf{Guangliang Liu\IU}
  ~~~~~\textbf{Xi Chen\NTU}
  ~~~~~\textbf{Bocheng Chen\UM}
  ~~~~~\textbf{Han Zi\IU}\\
  \textbf{Xitong Zhang\QC}
  ~~~~\textbf{Yuda Chen\UCSB}
  ~~~\textbf{Kristen Johnson\MSU}\\
  \IU Indiana University Indianapolis
  ~~\NTU Nanyang Technological University
  \UM University of Mississippi\\
  ~~~~~\QC Qualcomm
  ~~~~\UCSB University of California, Santa Barbara
  ~~~~~\MSU Michigan State University\\
\texttt{\{liugua,zihan\}@iu.edu}
  ~~~
\texttt{zoexi.chen@ntu.edu.sg}
  ~~~
\texttt{bchen5@olemiss.edu}\\
\texttt{zhangxit7@gmail.com}
~~~\texttt{ychen178@ucsb.edu}
~~~\texttt{kristenj@msu.edu}
}

\begin{document}
\maketitle
\begin{abstract}
Achieving social reasoning, including moral reasoning, cultural awareness, and politeness, is critical for the socially reliable deployment of large language models (LLMs). 
A key bottleneck is limited generalization: social situations that convey similar social implicatures can be expressed through highly diverse linguistic forms, creating an enormous space of possible situations and posing a substantial computational challenge even for LLMs. 
From a linguistic perspective, this challenge arises because social reasoning operates at the pragmatic level, requiring a pragmatic inference process to infer social implicatures from a given situation and its underlying social context.
Although moral values such as moral foundations are generally considered as the representations of the social context, how to construct the corresponding pragmatic inference steps remains underexplored.
Taking moral reasoning as an instance of social reasoning, this paper develops a pragmatic inference approach that enables LLMs to infer moral implicatures by (1) specifying the inferential steps through which LLMs derive moral implicatures from language, and (2) explicitly modeling the \textit{metapragmatic links} among these inferential steps to bridge the pragmatic gaps between them and make such inferences learnable for LLMs.
Experimental results demonstrate that our approach substantially improves LLMs' generalization in moral reasoning, highlighting the potential of pragmatic inference for future social reasoning research.
\textit{\footnotesize\textbf{Warning}: this paper contains  offensive language.}
\end{abstract}

\section{Introduction\label{sec:intro}}
With the rapid advancement of Large Language Models (LLMs), developing their capabilities for social reasoning, namely, enabling them to make socially aware decisions, has become an increasingly important research problem~\cite{gandhi2023understanding,strachan2024testing,shen2024understanding,yang_socially_2025}.
Among these social reasoning capabilities, moral reasoning is a central task.
Table~\ref{tab:moralreasoningexample} demonstrates an example of the elements that are involved in moral reasoning: moral situation, rule of thumb (RoT), moral foundation, and moral judgment. 
Given a situation, \bi{moral foundations}~\cite{haidt2004intuitive,graham2013moral} encode the underlying moral values (e.g., care or harm) that the situation either upholds or violates. 
A \bi{RoT} provides a situated explanation that guides the evaluation of the situation as right or wrong (e.g., murder is wrong)~\cite{forbes2020social}. 
The \bi{moral judgment} then follows by determining whether the situation is morally acceptable or unacceptable; in pragmatic terms, this judgment can be understood as a moral implicature.
In the current data, ``agree'' means the reply in the moral situation is acceptable, while ``disagree'' is the opposite. ``Neutral'' shows that the annotator neither agrees nor disagrees with the reply.
\begin{table}[t]
    \centering
    \begin{tabular}{l}
        \toprule
        \textbf{Moral Situation:} 
        Brush your teeth with\\laundry detergent.\\
        \midrule
        \textbf{Moral Foundation}: Care.\\
        \textbf{Rule of Thumb (RoT)}: It is harmful to brush\\teeth with laundry detergent.\\
        \textbf{Moral Judgment}: Disagree.\\
        \bottomrule
    \end{tabular}
    \caption{\small A Moral Reasoning Example. The Moral Foundations, Rule of Thumb and the Moral Judgment are all applied to the given reply. The Rule of Thumb (RoT) represents a judgment regarding an action mentioned in the prompt-reply, grounded in the moral foundation of care and sanctity. The moral situation consists of both the prompt and the corresponding reply.}
    \label{tab:moralreasoningexample}
\end{table}

Thus far, learning paradigms, which aim to equip LLMs with moral reasoning, generally train models to directly map a moral situation to moral judgment\footnote{Some studies also view RoT prediction as a task for evaluating moral reasoning.}, with or without a reference to moral foundations ~\cite{hendrycks2020aligning,forbes2020social,jiang2021can,ziems2022moral,jiang2025investigating}.
According to Rademacher complexity theory~\cite{li2025generalization,huang2025formal}, the overparameterized nature of LLMs should, in principle, make them highly effective at approximating the mapping from moral situations to moral judgments, as they have demonstrated strong function-approximation capabilities across many NLP tasks.
However, \bi{direct mapping suffers from limited generalization}~\cite{rao2023ethical,preniqi2024moralbert,jiang2025investigating}.

From a linguistic standpoint, this is mainly due to the pragmatic nature of morality~\cite{liu2025diagnosing,hendrycks2020aligning,tennant2024moral}.
In other words, moral situations that convey the same moral implicature, or judgment, can be expressed through highly diverse linguistic forms, creating an enormous space of situations that LLMs would need to fit in order to achieve satisfactory performance.
Empirical evidence from prior studies~\cite{ziems2022moral,rao2023ethical,jiang2025can} suggests that \textbf{\textit{the linguistic diversity of moral situations exceed the generalization capacity of current LLMs}}.

Because of the pragmatic nature of morality, moral judgment requires LLMs to infer (1) the social norms relevant to a given situation and (2) whether the situation conforms to or violates those norms.
This process is as \bi{pragmatic inference}\footnote{For the remainder of this paper, we use ``pragmatic inference'' to refer to the process of pragmatic inference.}: deriving the moral implicature, namely the moral judgment in moral reasoning, from a given situation and its underlying social context~\cite{hilton1995social,lenci2008distributional,purver2015distributional}.
Prior studies in moral reasoning either overlook the pragmatic inference~\cite{hendrycks2020aligning,tennant2024moral} or focus on benchmarking~\cite{ren-etal-2024-valuebench,yao2024value,kumar-jurgens-2025-rules} and measuring the moral values embedded in LLMs~\cite{scherrer2023evaluating,jiang2025can}, with limited exploration of approaches for pragmatic inference.

Although we have observed the effectiveness of pragmatic inference-driven approaches for other morality-related tasks, such as toxicity identification~\cite {sap2020social,zhou-etal-2023-cobra,chen2025pragmatic}. 
However, there are two challenges in applying pragmatic inference for moral reasoning. 
\textbf{First}, it remains unclear what inferential steps should constitute pragmatic inference for LLMs.
Motivated by findings from moral psychology, social science, and linguistics, we propose inferential steps centered on Moral Foundations Theory, which has been widely used in prior studies as a representation of social context~\cite{forbes2020social,ziems2022moral}.
Our proposed pragmatic approach regards moral foundations as the social context and consider a moral situation as ``disagree'' if the moral situation violates its underlying moral foundations.

\textbf{Second}, LLMs excel at capturing distributional semantics and generalizing through semantic similarity~\cite{liu-etal-2025-diagnosing}. 
However, the inferential steps required for pragmatics-level language understanding operate over social norms, intentions, consequences, and contextual assumptions, creating a \bi{pragmatic gap} between these steps and the distributional semantics encoded by LLMs.
To close this pragmatic gap, the concept of \textit{metapragmatic links}\footnote{A more detailed discussion is provided in Section~\ref{sec:relatedworks}. Notably, existing studies lack a principled methodology for constructing metapragmatic links in moral reasoning. We therefore adopt an empirical approach to demonstrate how these links facilitate pragmatic inference in LLMs.} motivates us to \textit{elaborate each inferential step by teaching LLMs not only what to infer, but also why such inferences are justified}.
The metapragmatic links perform as a bridge between semantic similarity and pragmatic inference~\cite{hubler2011metapragmatics}.

In this paper, we apply our pragmatic inference approach to three different moral reasoning tasks with different objectives, to demonstrate its adaptability and generalizability. 
The results shed light on a widely-applicable way for LLMs to identify moral issues from input and to regulate their own performance in accordance with human moral values.

\section{Related Works}\label{sec:relatedworks}
\begin{figure*}[ht]
    \centering
    \includegraphics[width=0.98\linewidth]{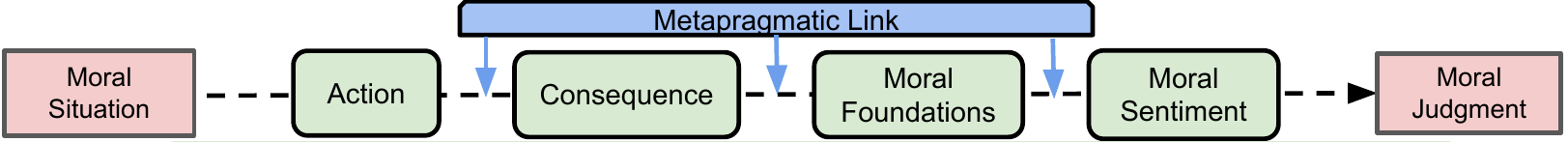}
    \caption{\textbf{Our Pragmatic Inference Approach for Moral Reasoning}.
It consists of four main steps: (1) \textbf{Action}: identifying the actions involved in the moral situation; (2) \textbf{Consequence}: estimating the consequences of those actions; (3) \textbf{Moral Foundations}: predicting the relevant moral foundations associated with the actions and determining whether those actions violate or uphold the identified foundations; and (4) \textbf{Moral Sentiment}: identifying the sentiment of the situation with respect to the identified actions. 
} 
    \label{fig:inferencechain}
\end{figure*}

\textbf{Pragmatic Inference and Metapragmatics}. 
It is widely known that distributional semantics is one of the theoretical foundations for embeddings. It interprets the meaning of a linguistic unit by additional signs (e.g., words, emoji) surrounding it~\cite{harris1954distributional,firth1957synopsis,mcdonald2001testing,lenci2023distributional,boleda2020distributional}. In other words, it treats the ``context'' as the texts provided, neglecting the many ``unsaid'' context of social variables that underlie real interactions, such as the social constraints that a speaker encounters, the effects that a speech creates on the listeners, and the delivery of intention~\cite{lenci2008distributional,purver2015distributional}. These are studied in pragmatics -- the linguistic discipline that studies language in context. Possibly due to the above-mentioned gap between ``having internal representations of social variables'' and ``drawing effectively on them'' \cite{yang_socially_2025},  LLMs have shown several pragmatic deficiencies, e.g., understanding implicature \cite{cong_manner_2024}, comprehending metaphors and humours \cite{ruis_goldilocks_2023, barattieri_di_sanpietro_pragmatic_2023}, and moral alignment~\cite{khamassi2024strong,liu2024towards}.

In this work, we leverage pragmatic inference to facilitate LLMs to achieve stronger performance in moral reasoning. To reiterate, pragmatic inference refers to the process of deriving conclusions of meanings based on context, intention, and language use \cite{elder_pragmatic_2024}. 
The inferential process is, however, not often explicitly stated even in human interactions. It is rather common from human speakers to present their moral judgment without specifying how they make it. The inferences thus stay non-verbally at the level of metapragmatic awareness, where meta-links are built between an instance of language use and social variables (e.g., linking the phrase ``thank you'' to the social norm of politeness) \cite{kadar_understanding_2013}.  
Humans learn these metapragmatic links not just from language input, but also through physical and social experiences. The latter is absent in LLMs. Therefore, for LLMs to learn, the inferential process where metapragmatic links are built needs to be textualized, involving relevant social variables, such as context and moral foundations. 

\textbf{Moral Foundations Theory (MFT).} \citet{haidt2004intuitive} and \citet{graham2013moral} describe a framework for understanding human morals and moral reasoning. 
MFT hypothesizes that moral judgments are guided by domain-specific foundations, i.e., care, fairness, loyalty, liberty, authority, and sanctity. 
These foundations capture the diversity of moral concerns and have been widely adopted in computational social science~\cite{johnson2018classification,wendell2021classifying,reinig2024survey}. 
They provide a principled way to annotate, analyze, and model moral dimensions in language, making them particularly relevant for studying how LLMs represent and reason about morality.

\section{Methodology\label{sec:method}}

In this section, we first introduce our pragmatic inference approach and then describe how it can be adapted to different moral reasoning objectives, namely moral foundations classification (MFC) and moral judgment. Our pragmatic inference approach is also flexible with different types of input. In the current study, we tested moral situation or RoT as input\footnote{We note that pragmatic inference often admits multiple pathways to the same conclusion, and our approach represents one such pathway.}. 

\begin{figure*}[ht]
    \centering
    \includegraphics[width=1.0\linewidth]{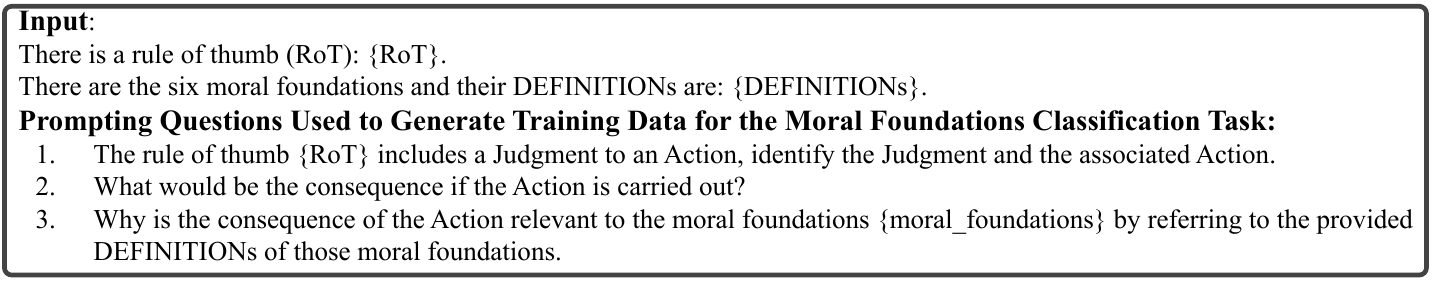}
    \caption{\textbf{Prompting Off-the-Shelf LLMs to Generate \textit{Training Data} for Moral Foundations Classification}. Those prompting questions are designed according to our proposed pragmatic inference approach shown in Figure~\ref{fig:inferencechain}. All content within \{ \} is provided by the benchmark. We adopt the MIC benchmark~\cite{ziems2022moral}, which consists of prompt–reply pairs, where each pair is treated as a moral situation and the moral reasoning objectives are defined with respect to the reply.}
    \label{fig:inference4MFC}
\end{figure*}

\subsection{Our Pragmatic Inference Approach\label{sec:motivation}}

Proposing a pragmatic inference approach for moral reasoning is challenging because there is no linguistic theory that clearly defines what constitutes an appropriate pragmatic inference for LLMs~\cite{bender-koller-2020-climbing,mahowald2024dissociating}. 
To have those inferential steps, the theoretical frameworks of how humans make pragmatic inferences and moral judgments are borrowed. However, this study does not assume their direct applicability to LLMs. Instead, \textit{this paper proposes a pragmatic inference approach after careful examinations with LLMs}.

Figure~\ref{fig:inferencechain} introduces the design and its rationales: 
The \textbf{Action} component is to identify mentioned actions in the given moral situation. According to the famous pragmatics theory of speech act, humans speak not for the purpose of speaking, but for doing things \cite{austin_how_1975}. Thus, the moral judgment is often, if not always, a judgment of the action that the speaker is expressing (e.g., ``I want to destroy the world'') or that others once performed (e.g., ``who said he is an idiot'')~\cite{forbes2020social,ziems2022moral}. We set ``Action'' as the first step, foregrounding the rest of steps.

The second step \textbf{Consequence} is to recognize the consequence of carrying out the ``Action''. Moral judgment is, in nature, an evaluation of good/bad placed on the outcome of an action~\cite{nozick1968moral,schaich2006consequences,emelin2021moral}.
The third step \textbf{Moral Foundations} (MFs) then moves from the individual action/consequence to the \bi{social context} that underlies them. The effectiveness of including MFs in moral reasoning has been supported by previous theories and experiments ~\cite{graham2013moral,abdulhai2024moral}. 
Given the results of Moral Foundations step (e.g., an action violates a moral foundation), the last step \textbf{Moral Sentiment} is necessary to unravel the relationship between actions, consequences, and the moral situation. 
For example, if an action is moral and the situation shows a positive sentiment about it, the situation is likely moral.
The importance of moral sentiment in causally shaping moral judgment has been shed light on by the Social Intuitionist Model~\cite{haidt2001emotional,zollo2021consumers}.

\bi{These inferential steps are connected through metapragmatic links, making them more explicit and learnable for LLMs grounded in distributional semantics.}
The connection between an action and its consequence requires the metapragmatic evaluation of social norms, e.g., the connection between the action of ``encouraging war'' and the consequence of ``civilian casualty'' is built on the oft-unspoken inference about how wars cause mass death. Similarly, attributing the consequence of an action to the abstract MFs demands the metapragmatic evaluation of how MFs guide and are also reflected in human behaviours. 
\textit{Through reifying the metapragmatic links between different inferential steps, this study teaches LLMs to effectively conduct moral reasoning}.

In the following subsections, we apply different parts of the framework to three moral reasoning tasks that have different objectives, to show its adaptability. Specifically, we first designed prompting questions according to the framework and used them to obtain training data from off-the-shelf LLMs. The training data is then used to train several small LLMs.  

\begin{figure*}[ht]
    \centering
    \includegraphics[width=1.0\linewidth]{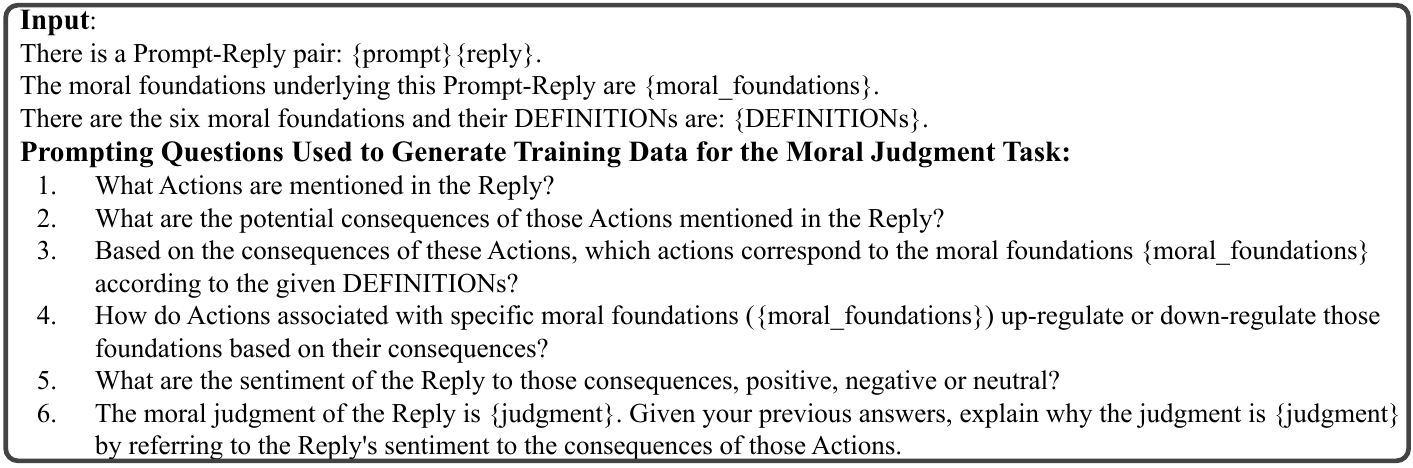}
    \caption{\textbf{Prompting Off-the-Shelf LLMs to Generate \textit{Training Data} for Moral Judgment based on Ground-truth Moral Foundations}. Those prompting questions are designed according to our proposed pragmatic inference approach shown in Figure~\ref{fig:inferencechain}. All content within \{\} is provided by the benchmark. We adopt the MIC benchmark~\cite{ziems2022moral}, which consists of prompt–reply pairs, where each pair is treated as a moral situation and the moral reasoning objectives are defined with respect to the reply.}
    \label{fig:inference4Judgment}
\end{figure*}

\subsection{Task 1: Moral Foundations Classification (MFC)}

The first task demonstrates how to identify the moral foundations from a given RoT. It requires only a light load of pragmatic inference, because RoT is a situated explanation for the good or bad nature of a moral situation, in other words, it already has a judgment on the given situation, e.g., ``murder is wrong''. Figure~\ref{fig:inference4MFC} presents the prompting questions of the inferential steps used in the MFC task.
The input here is a simple RoT and the definitions of six MFs. The inferential steps are merely the first three components of our pragmatic inference approach (Figure~\ref{fig:inferencechain}), namely, action, consequence, and MFs.



\subsection{Task 2: Moral Judgment}
Figure~\ref{fig:inference4Judgment} shows the prompting questions used in the second task. 
Unlike the first MFC task, where RoTs are presented without context, the input here is a moral situation consisting of a prompt–reply pair along with the associated ground-truth MFs.
In other words, we provide the specific context (i.e., prompt and reply) and test whether LLMs can effectively make context-specific moral judgments based on MFs.


The second task include all components of our framework, namely, action, consequence, MFs, and moral sentiment, given the increasing complexity of its pragmatic inference. Especially, the MFs component is divided into two steps: step 3 relates action to the ground-truth MFs, while step 4 further specifies whether the relation is a violation or observation of the MFs\footnote{Please note that the empirical results indicate that applying Step 3 or Step 4 individually performs better than combining them.
This is reasonable, as each step constructs a metapragmatic link, and applying the steps individually makes this link easier to establish.}.
We should note that, during the training, judgment is provided for LLMs to learn its connections with the other components (step 6). In the actual test, LLMs are required to produce the judgment by themselves. 


\subsection{Task 3: MFC+Moral judgment\label{sec:mfijudg}}
Building on the first two tasks, the third task teaches LLMs to jointly infer MFs from a given \textit{situation} and make corresponding moral judgments, namely, MFC + moral judgment.
For the prompting questions, we drop the given ground-truth moral foundations in the input and replace the step 3 and 4 in Figure~\ref{fig:inference4Judgment} with the following prompt questions:
\begin{tcolorbox}[
  colback=black!1!white,      
  colframe=black!80,          
  boxrule=1.5pt,            
  arc=1mm,       
  left=2mm, right=2mm, top=1mm, bottom=1mm,
  fontupper=\small, 
  breakable           
]
(3). Based on the consequences of these Actions, why are their underlying moral foundations classified as \{moral\_foundations\} according to the provided DEFINITIONs?\\
(4). How do Actions up-regulate or down-regulate their underlying moral foundations based on their consequences?
\end{tcolorbox}

In doing so, \textit{we guide LLMs to learn to infer the underlying moral foundations}, motivated by their strong performance on the MFC task (Figure~\ref{fig:inference4MFC}).
The input is kept to the situation (Prompt+Reply) and the definitions of MFs only, no ground-truth MFs are provided. This setting is closer to the real life where a speaker only hears the interaction and has some (implicit) knowledge of MFs. 

\section{Experimental Setting and Results\label{sec:preliminary}}

In this section, we present the experimental settings and results to evaluate the effectiveness of our pragmatic inference approach for moral reasoning. 
Further analyses of our approach are provided in Appendix~\ref{app:further_analysis}, including evidence that our pragmatic inference approach does not rely on shallow heuristics but heavily relies on moral foundations.
\subsection{Experimental Settings}
\paragraph{Benchmark.}
In this paper, we adopt the Moral Integrity Corpus (MIC) benchmark and its annotations from~\citet{ziems2022moral}\footnote{Artifact Licensing. We use publicly available datasets and pretrained models in accordance with their respective licenses and terms of use.}. We chose the MIC benchmark because it closely reflects how humans make moral judgments in daily communication and provides richer contextual information.
One advantage of the MIC dataset is that it provides levels of agreement between the annotators, which allows us to focus on the less controversial subset of the benchmark. Indeed, we adopt 26\% of the samples that have received full agreement among all annotators, given that moral judgments are subjective and accounting for subjectivity is beyond the current research scope.
The MIC benchmark utilizes the six MFs defined by~\citet{graham2013moral}: care, fairness, liberty, loyalty, authority, and sanctity. In the data, up to three MFs could be associated with one moral situation. 
As presented above, we utilize two of the evaluations as our objectives: MFs and moral judgment (agree, neutral, or disagree with the reply). 
RoT is not selected as a learning objective because: (i) it does not have a fixed answer and, more crucially, multiple RoTs may exist in one situation\footnote{We refer readers to Section 3.1 of~\citet{ziems2022moral} for further details.}, and (ii) it has a rather dynamic relationship with MFs.

\begin{table}[ht]
\centering
\setlength{\tabcolsep}{3pt} 
\small
\begin{tabular}{@{}cccc|c@{}}
\toprule
\multicolumn{4}{c|}{\textbf{MFC}} & \multirow{2}{*}{\textbf{\makecell{Moral\\Judgment}}} \\
\cmidrule(lr){1-4}
\textbf{\#MFs=1} & \textbf{\#MFs=2} & \textbf{\#MFs=3} & \textbf{Average} & \\
\midrule
.694 & .274 & .147 & .579 & .466\\
\bottomrule
\end{tabular}
\caption{Performance of Direct Prompting Off-the-shelf LLM (deepseek-chat). The prompts used in these evaluations are provided in Appendix \ref{fig:directprompt}.}
\label{tab:directprompting} 
\end{table}

To examine how moral reasoning performance scales with the size of the training corpus, we consider three data settings.
We use subsets of 5K, 10K, and all of the 23500 samples of the full-agreement data from the training set as the fine-tuning data.
We report all performance results using a fixed development set, with models trained on different subsets of the full-agreement training data.
Therefore we can verify how the training data scales can influence the moral reasoning performance.
To ensure the selected dataset from this 2022 study is still appropriate for the current study, we first prompt the deepseek-chat model directly. Table \ref{tab:directprompting} presents its performance on both moral foundation prediction and moral judgments. They clearly show that moral reasoning remains a challenging task even for advanced LLMs.

Therefore, following the steps of our pragmatic inference methods (detailed in Section~\ref{sec:method}), we prompt the DeepSeek model to provide responses to each of the inferential components as well as their metapragmatic links. 
The step-wise answers are then used as the training data\footnote{The data used in our experiments is available in the anonymized GitHub repository: \url{https://anonymous.4open.science/r/moralpragmatics-F891}.}  for fine-tuning the LLMs towards moral reasoning objectives. 

 \begin{table*}[ht]
\centering
\setlength{\tabcolsep}{2.5pt} 
\begin{tabular}{@{}ll | cccc |cccc| cccc@{}}
\toprule
& & \multicolumn{4}{c}{\textbf{Accuracy(\#MFs=1)}} & \multicolumn{4}{c}{\textbf{Accuracy(\#MFs=2)}} & \multicolumn{4}{c}{\textbf{Accuracy(\#MFs=3)}} \\
\cmidrule(lr){3-6} \cmidrule(lr){7-10} \cmidrule(lr){11-14}
\textbf{Model} & \makecell{\textbf{Data Scale}} 
& \textbf{base} & \textbf{base+} & \textbf{CoT} & \textbf{ours} 
& \textbf{base} & \textbf{base+} & \textbf{CoT} & \textbf{ours} 
& \textbf{base} & \textbf{base+} & \textbf{CoT} & \textbf{ours} \\
\midrule
\multirow{3}{*}{\shortstack[l]{Llama-1B}} 
& 5000    & .501 & .696 & .661 & \textbf{.890} & .402 & .545 & .475 & \textbf{.856} & .396 & .460 & .428 & \textbf{.806} \\
& 10000   & .582 & .593 & .677 & \textbf{.832} & .489 & .449 & .545 & \textbf{.778} & .414 & .365 & .468 & \textbf{.743} \\
& 235000 & .552 & .671 & .717 & \textbf{.770} & .505 & .514 & .509 & \textbf{.796} & .387 & .495 & .401 & \textbf{.743} \\
\midrule
\multirow{3}{*}{\shortstack[l]{Llama-3B}} 
& 5000    & .537 & .754 & .719 & \textbf{.851} & .466 & .653 & .544 & \textbf{.822} & .423 & .572 & .459 & \textbf{.798} \\
& 10000   & .386 & .558 & .700 & \textbf{.725} & .336 & .507 & .507 & \textbf{.649} & .324 & .451 & .441 & \textbf{.698} \\
& 235000 & .653 & .790 & .727 & \textbf{.839} & .493 & .645 & .533 & \textbf{.784} & .451 & .617 & .414 & \textbf{.743} \\
\midrule
\multirow{3}{*}{\shortstack[l]{Mistral-7B}} 
& 5000    & .612 & .649 & .677 & \textbf{.803} & .474 & .504 & .533 & \textbf{.634} & .441 & .469 & .401 & \textbf{.577} \\
& 10000   & .456 & .653 & .653 & \textbf{.754} & .373 & .466 & .515 & \textbf{.606} & .401 & .419 & .523 & \textbf{.536} \\
& 235000 & .487 & .669 & .649 & \textbf{.768} & .442 & .536 & .420 & \textbf{.604} & .383 & .446 & .383 & \textbf{.559} \\
\bottomrule
\end{tabular}
\caption{\small\textbf{Moral Foundations Classification (MFC)} Performance Across Different Training Dataset Scales. The majority of prompt-reply pairs involve no more than three moral foundations labels; therefore, we report the performance on cases containing fewer than four moral foundations. For example, \textbf{\#MFs=2} indicates performance on instances annotated with a label containing two moral foundations. The optimal performance is highlighted with a \textbf{bold} font.}
\label{tab:MFC_performance}
\end{table*}

\paragraph{Backbone Models.} We chose Llama3-3.2-1B, Llama3-3.2-3B~\cite{grattafiori2024llama3herdmodels}  and Mistral-7B (Mistral-7B-v0.3~\cite{jiang2023mistral7b}) as our backbone models, and leverage the supervised fine-tuning objective to train those models by following previous studies~\cite{ziems2022moral,liu2025diagnosing}.
We intentionally select smaller models, as they are generally considered less capable than their larger counterparts. 
We aim to demonstrate the effectiveness of our pragmatic inference approach through the strong performance achieved by these smaller models.

\paragraph{Evaluation.}
To reiterate, we have three moral reasoning tasks. Prediction accuracy is used as the evaluation metric across them. In the meantime, the prediction accuracy of MFs is further divided by the data that refers to one, two, or three MFs, respectively.
In other words, we have taken into consideration the one-to-many relationship between RoT/situation and MFs.

\paragraph{Method Setting and Baseline Methods.}
We denote the LLM $f$ is parameterized by $\theta$, and the definitions (see Figure~\ref{fig:mftdef}) of MFs are represented with $d_m$.
Let $(x_S, x_R, y_M, y_J)$ denote a sample from the MIC benchmark, where $x_S, x_R$ represents the moral situation and RoT, respectively. $y_M$ is its associated MF label. Each $y_M$ may contain between one and three MFs. $y_J$ is the moral judgment.
We use $|y_M| = i$ $(1 \leq i \leq 6, x \in \mathbb{Z})$ to indicate that $x$ is associated with $i$ MFs.
We compute accuracy for each value of $i$.

For \textbf{baseline} methods, we follow prior work~\cite{forbes2020social,ziems2022moral} and consider two settings: 
\textbf{(1)} The \textbf{base} setting. For the moral judgment tasks, we compute $y_J = f_{\theta}(x_S)$; and for MFC, we compute $y_M = f_{\theta}(x_R)$. 
\textbf{(2)} The \textbf{base+} setting.
For the moral judgment task, we compute $y_J = f_{\theta}(x_S, y_M)$ wherein the ground-truth MFs are provided; and for MFC task, we compute $y_M = f_{\theta}(x_R, d_m)$ wherein only the definition of six MFs are provided. 
\textbf{(3)} The \textbf{CoT} setting. We also consider a strong baseline based on the standard Chain-of-Thought (CoT) method~\cite{wei2022chain}. 

CoT assists LLMs to divide a complex task into intermediate steps and has been proven to be effective in logical reasoning tasks \cite{fang_cdw-cot_2025, ji_mygo_2025}. However, it has also been identified as less effective in non-logical reasoning tasks, such as pragmatic inference \cite{sprague_cot_2024, chen2025pragmatic}. 
To ensure that the effectiveness of CoT is not overlooked, we apply the same training process, namely, obtaining training data by using CoT to prompt the same off-the-shelf LLMs\footnote{Please refer to Appendix~\ref{fig:cotprompt} for the details of the prompting questions.} and then use the step-wise answers to train the models. We compute  $y_J = f_{\theta}(x_S, \{\text{CoT\_inference}\})$ and $y_M = f_{\theta}(x_R, \{\text{CoT\_inference}\})$ for moral judgment and MFC, respectively.

For our pragmatic inference approach, we compute $y_M = f_{\theta}(x_R, d_m, \{\text{ours\_inference}\})$ for Task 1, denoted as \textbf{ours-MFC}.
For Tasks 2 and 3, we denote 
$
y_J = f_{\theta}(x_S, y_M, \{\text{ours\_inference}\})
$
as \textbf{ours-MJ}, where moral judgment (MJ) is the objective; and we denote 
$
y_J, y_M = f_{\theta}(x_S, \{\text{ours\_inference}\})
$
as \textbf{ours-joint}, where the LLMs are asked to predict both MFs and the moral judgment.
More details about hyperparameters and experimental setup are in Appendix~\ref{app:experiment}.

\subsection{Experimental Results}

\begin{table}[ht]
\centering
\small
\setlength{\tabcolsep}{6pt}
\begin{tabular}{ll | cccc}
\toprule
\textbf{Model} & \textbf{\makecell{Data\\Scale}} 
& \textbf{base} & \textbf{base+} & \textbf{CoT} & \textbf{\makecell{ours\\-MFC}} \\
\midrule
\multirow{3}{*}{\shortstack[l]{Llama\\1B}}
& 5000    & .433 & .567 & .521 & \textbf{.851} \\
& 10000   & .495 & .469 & .564 & \textbf{.784} \\
& 235000 & .481 & .560 & .542 & \textbf{.770} \\
\midrule
\multirow{3}{*}{\shortstack[l]{Llama\\3B}}
& 5000    & .475 & .659 & .574 & \textbf{.824} \\
& 10000   & .349 & .505 & .549 & \textbf{.691} \\
& 235000 & .532 & .684 & .558 & \textbf{.789} \\
\midrule
\multirow{3}{*}{\shortstack[l]{Mistral\\7B}}
& 5000    & .509 & .540 & .537 & \textbf{.671} \\
& 10000   & .410 & .513 & .563 & \textbf{.632} \\
& 235000 & .437 & .550 & .484 & \textbf{.643} \\
\bottomrule
\end{tabular}
\caption{\small\textbf{Average Moral Foundations Classification (MFC)} Performance Across Different Training Dataset Scales. Our proposed pragmatic inference methods outperform all baseline methods and the CoT method.}
\label{tab:average_MFC_performance}
\end{table}
\paragraph{Task 1: MFC.} Table~\ref{tab:MFC_performance} reports the models' performance on the development set of data that have 1–3 moral foundation labels. 
Due to the space limit, we report the average performance for each experimental setting in Table~\ref{tab:average_MFC_performance}. 
Across all training dataset sizes, model variants, and numbers of moral foundation labels, our method consistently outperforms the baseline methods and the CoT method.
Comparing the results of \textbf{base} and \textbf{base+}, it is evident that incorporating the definition of MFs is beneficial, which aligns with findings from previous studies~\cite{ziems2022moral,liu2025diagnosing}.
Our method also consistently outperforms the \textbf{CoT} method, which is, at times, worse than the \textbf{base+} method. 

This finding corroborates previous observations of the inefficiency of CoT in pragmatic inference \cite{chen2025pragmatic, sprague_cot_2024}. Another noteworthy finding is that, unlike the baseline methods whose performance drops significantly as the number of underlying MFs increases, our method exhibits much smaller performance degradation. 
Interestingly, neither increasing the model size nor increasing the training dataset leads to stable performance gains. 
The possible reasons include that the additional data does not substantially expand the range of unique moral situations, and 
the same size of training data is insufficient to modify the knowledge that larger LLMs have already acquired during pre-training compared to smaller models.

\begin{table}[ht]
\centering

\small %
\setlength{\tabcolsep}{4pt} 
\begin{tabular}{@{}ll|ccc|cc@{}}
\toprule
\textbf{Model} & \textbf{Size} & \textbf{base} & \textbf{base+} & \textbf{CoT} & \textbf{\makecell{ours-\\MJ}} & \textbf{\makecell{ours-\\joint}}\\
\midrule


\multirow{3}{*}{\makecell{Llama\\-1B}} 
& 5000  & .151 & .080 & .629&\textbf{.708} &.706\\
& 10000 & .498 & .335 & .652& \textbf{.728} &.718\\
& 23500 & .179 & .318 & .641&\textbf{.710} &.704\\
\midrule

\multirow{3}{*}{\makecell{Llama\\-3B}} 
& 5000  & .160 & .304 & .635& .712 &\textbf{.750} \\
& 10000 & .205 & .252 & .653&.735 &\textbf{.748}\\
& 23500 & .249 & .390 & .665&.710 &\textbf{.766}\\
 \midrule

\multirow{3}{*}{\makecell{Mistral\\-7B}} 
&5000& .418 &.366 & .637&\textbf{.704} &.694\\
&10000&.438&.434&.628&\textbf{.760} &720\\
&23500&.486&.358&.644&.696 &\textbf{.704}\\

\bottomrule
\end{tabular}
\caption{\small Accuracy of the Moral Judgment Task Across Training Dataset and Model Scales. The optimal performance is highlighted with a \textbf{bold} font. Our proposed pragmatic inference approach, i.e., \textbf{ours-MFs} and \textbf{ours-joint}, outperform all baseline methods and the CoT method. Although \textbf{ours-joint} infers the MFs, its performance is comparable to \textbf{ours-MFs}, which has access to the ground-truth MFs.}
\label{tab:performance4judge}
\end{table}
\paragraph{Tasks 2 and 3: Moral Judgment.} Table~\ref{tab:performance4judge} presents the performance on the moral judgment tasks. 
Again, our pragmatic inference methods (\textbf{ours-MJ} and \textbf{ours-joint}) consistently outperform both baseline methods (\textbf{base, base+}) and the \textbf{CoT} method across different training dataset scales and model architectures. Similar to the MFC task, increasing data size does not necessarily improve the model performance.
Since moral judgment is a three-class classification task (agree, disagree, or neutral), where random guessing yields an accuracy of approximately 0.333, the baseline methods perform worse than random guessing in some settings of dataset and model sizes. 

In contrast, baseline methods perform much better than random guessing on the MFC task. The difference corroborates that the pragmatic gap varies in size between the two tasks. This is understandable as the involvement of context (prompt+reply) in the second task brings about more variability in interpretations than the close alignment of RoT and normative interpretations, for example, the RoT ~``\textit{It is good to keep in touch with friends}'' can rather be easily attributed to the relational norms between friends. It is, however, worth emphasizing that our methods have been consistently effective across the different tasks, showing the adaptability and stability of our pragmatic inference approach. 

Noteably, CoT in tasks 2 and 3, although still falling behind our methods, outperforms the baselines. The finding suggests that the baseline designs are particularly weak in addressing complex tasks, which the step-wise nature of CoT provides an advantage. However, CoT as a trick designed primarily for logical reasoning tasks still has a rather restricted performance in pragmatic inference tasks. 

Recall that the difference between Task 2 and Task 3 is that Task 2 has the input of ground-truth MFs, whereas Task 3 needs to predict both MFs and moral judgment. Possibly with the involvement of ground-truth MFs, \textbf{ours-MJ} outperforms \textbf{ours-joint} in two of the three models. This does not deny the effectiveness of the design of \textbf{ours-joint}. In fact, it outperforms \textbf{ours-MJ} with Llama-3B, shedding light on the potential of LLMs automatically inferring both underlying MFs and making moral judgments. 
Please refer to Appendix~\ref{sec:discussion} for further discussion of our findings.

\section{Conclusion\label{sec:conclusion}}

In this paper, we propose a pragmatic inference approach to improve the generalization of social reasoning in LLMs, taking moral reasoning as the central case study.
Rather than relying on direct mappings from moral situations to moral judgments, we propose a pragmatic inference approach that treats moral foundations as the social context underlying moral situations and connects moral situations, social norms, and moral implicatures. To make these inferences learnable for LLMs, our approach explicitly models the \textit{metapragmatic links} among inferential steps by textualizing not only what should be inferred, but also why each inference is justified.

We evaluate this approach across three moral reasoning tasks with different objectives, demonstrating its flexibility and adaptability across distinct forms of moral reasoning objectives. 
Experimental results show that our approach consistently improves moral reasoning performance, reduces perplexity compared with direct-mapping baselines, and weakens reliance on shallow heuristics. 
These findings suggest that explicitly modeling pragmatic inference provides an effective pathway for helping LLMs better leverage their existing social knowledge and achieve more robust, generalizable moral reasoning.

\section*{Limitations}
Although the proposed pragmatic inference chains are highly effective overall, we do not evaluate the contribution of each individual step because of space constraints and because such a step-level analysis falls outside the scope of this paper. We thus encourage future studies to conduct further ablation experiments, as well as explore their variations. These future attempts may also extend the current experimental settings to other downstream morality-relevant tasks, such as jailbreaks. It is anticipated that the generalisability of our design will be further confirmed across the tasks that have different semantic distributions.

\section*{Ethics Statement}
This work uses datasets containing toxic language, social biases, and morally inappropriate prompts, which are necessary for training and evaluating models on the moral reasoning task. All datasets used in our experiments are publicly available and were originally collected for research purposes. To reduce the risk of misuse, we release only the model weights and evaluation code and do not redistribute raw fine-tuning data containing harmful content.
\bibliography{custom}

\appendix

\section{Appendix}
\label{sec:appendix}


\subsection{Experimental Details\label{app:experiment}}
We apply supervised fine-tuning with a learning rate of 5e-5 and a batch size of 24. 
For mistral-7B, we leverage the LoRA setting with rank: 64, lora alpha: 16, lora dropout: 0.1.
All experiments are conducted for 5 epochs using three random seeds (1, 2, and 3), and we report the optimal result across three random seeds.
~\textbf{Template for fine-tuning LLMs} is:
\textbf{base}: The rule-of-thumb judgment is \{RoT\}. The moral foundations underlying the rule-of-thumb are \{moral foundations\}.
\textbf{base+}: \{Definition of moral foundations\} The rule-of-thumb judgment is \{RoT\}. The moral foundations underlying the rule-of-thumb are \{moral foundations\}.
\textbf{ours}: \{Definition of moral foundations\} The rule-of-thumb judgment is \{RoT\}. \#\#\#Inference: \{inference\} the moral foundations underlying the rule-of-thumb are \{moral foundations\}.
\begin{figure}[h]
\small
    \centering
    \includegraphics[width=0.7\linewidth]{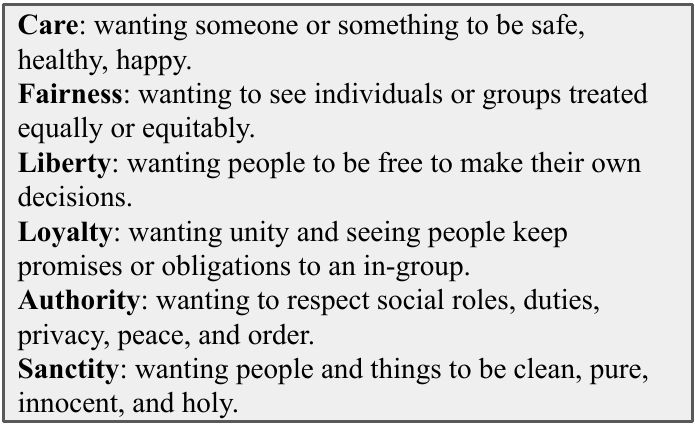}
    \caption{Definition of Six Moral Foundations}
    \label{fig:mftdef}
\end{figure}

\begin{figure}[h]
\small
    \centering
    \includegraphics[width=0.7\linewidth]{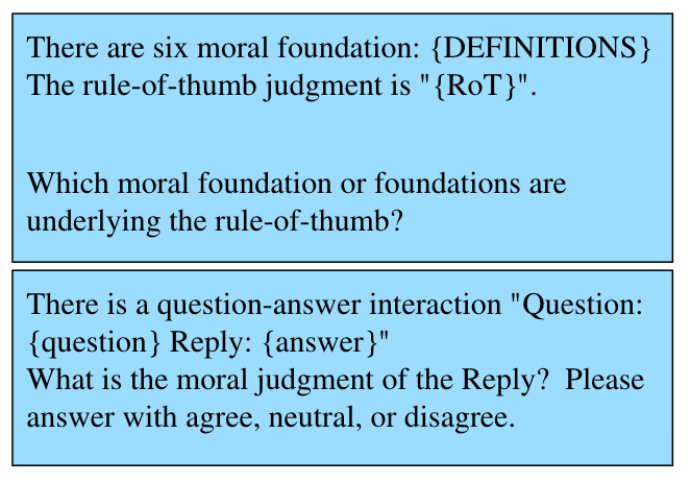}
    \caption{Prompts for Direct LLMs Prompting}
    \label{fig:directprompt}
    \vspace{-15pt}
\end{figure}

\begin{figure}[h]
\small
    \centering
    \includegraphics[width=0.7\linewidth]{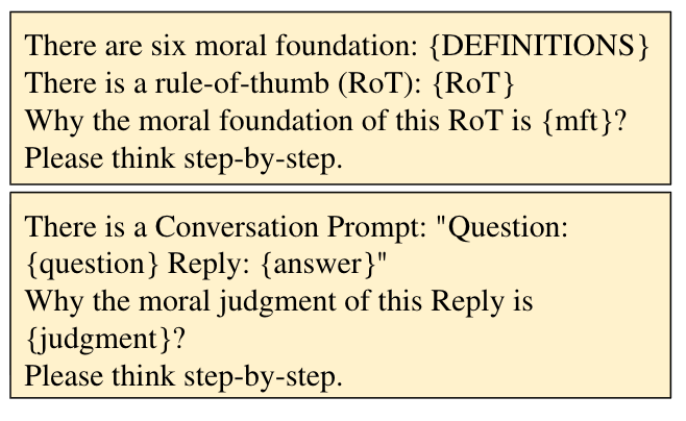}
    \caption{Prompts for Standard CoT Prompting}
    \label{fig:cotprompt}
\end{figure}

\subsection{Further Analysis\label{app:further_analysis}}
In the previous sections, we demonstrated the effectiveness of our proposed pragmatic inference approach. In this section, we present additional analyses to further characterize it.
Specifically, we: (1) conduct intervention experiments to examine whether our approach guides LLMs to avoid heuristics and utilizes MFs when making moral judgments; 
(2) calculate the perplexity to show that our approach makes moral reasoning discourse more learnable than baseline methods that overlook the pragmatic gap; 
and (3) perform a detailed analysis of individual MFs, showing that while performance on the majority of MFs is strong, dataset imbalance remains a key bottleneck to achieving good performance on rare MFs.
\textit{Notably, for all analytical experiments on the moral judgment task, we employ the design of Task 3, namely, jointly infering moral foundations and moral judgments.}

\paragraph{Dependence on Moral Foundations.}
Prior studies show that LLMs may not pragmatically leverage the cues of MFs or other morality-related explanations. Instead, they rely on superficial correlations to achieve generalization in pragmatics-level tasks~\citep{shapira2024clever,liu2025discourse,sanchez2025metaphor}.
Just as ~\citet{liu2025discourse} demonstrates, LLMs correct immoral outputs to moral ones, while neglecting the underlying moral issues provided to them. This gives rise to the issue of generalizability across different datasets. To test whether the same problem occurs in the current study, we conduct an intervention experiment to examine whether our pragmatic inference method has facilitated LLMs' effective use of MFs.

Specifically, we replace the automatically inferred MFs with the ground-truth MFs and observe how the performance changes. Note that this design is different from Task 2, because the ground truth MFs are not an initial input here. Instead, we let the LLMs complete their inference process with the predicted MFs, and then replace them with the ground truth MFs only in the answer. After that, LLMs are led to output their moral judgment. In other words, we pause the task right before the last step and do the replacement, then we check whether the 'airdropped' ground-truth MFs lead to better performance of LLMs, with everything else in the inference remaining the same.

\begin{figure}[h!]
    \centering
    \begin{minipage}[b]{0.235\textwidth}
        \includegraphics[width=\textwidth]{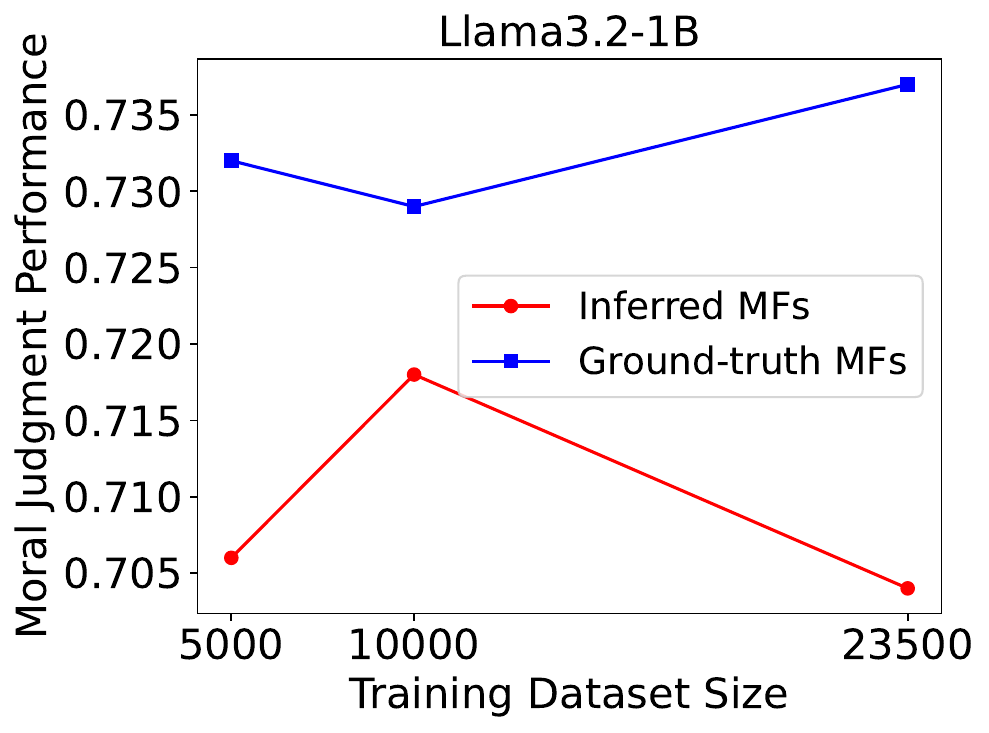}
    \end{minipage}\hfill
    \begin{minipage}[b]{0.235\textwidth}
        \includegraphics[width=\textwidth]{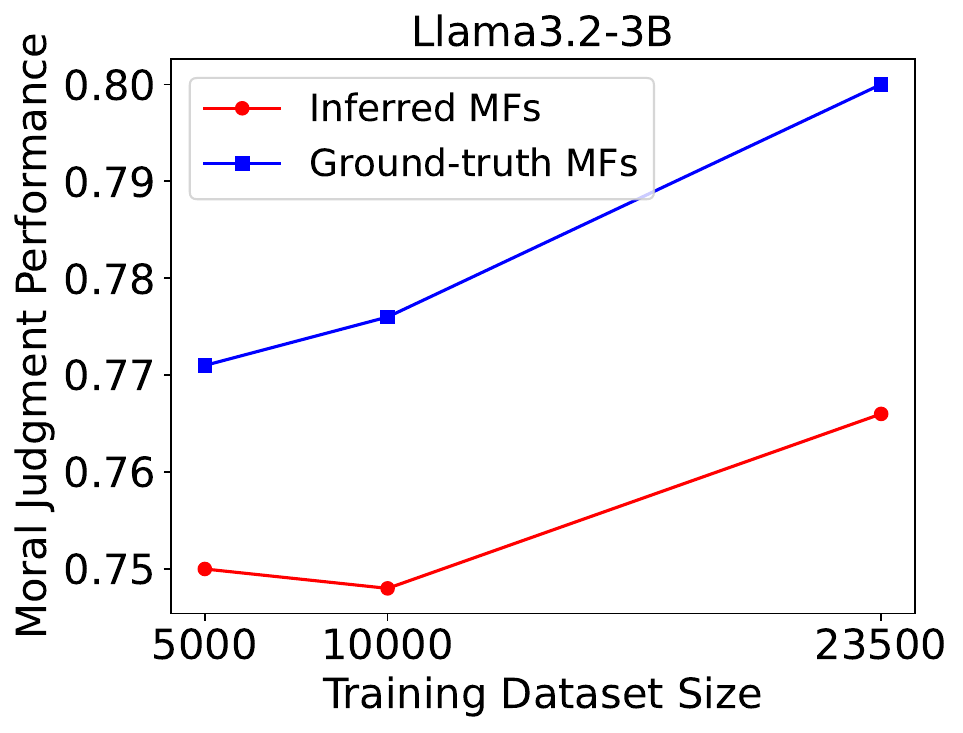}
    \end{minipage}
    \caption{Accuracy of the Moral Judgment Task  With Our Intervention Experiment. }
    \label{fig:heuristics}
\end{figure}

Figure~\ref{fig:heuristics} exhibits that the replacement with ground-truth MFs consistently improves performance compared to the experiment without this intervention.
In other words, the current pragmatic inference method has indeed facilitated LLMs to make effective use of MFs. Given MFs are commonly grounded in various moral situations, the finding suggests the potential for the current method to be generalized to other morality-related tasks.

\begin{figure}[h!]
    \centering
    \begin{minipage}[b]{0.240\textwidth}
        \includegraphics[width=\textwidth]{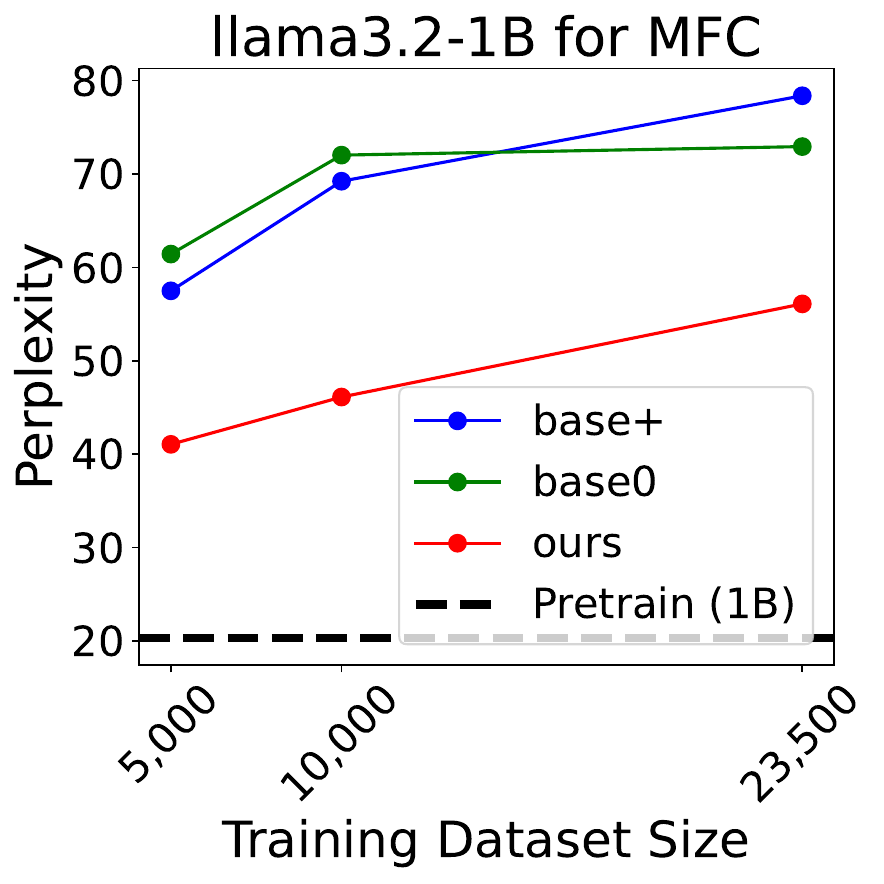}
    \end{minipage}\hfill
    \begin{minipage}[b]{0.240\textwidth}
        \includegraphics[width=\textwidth]{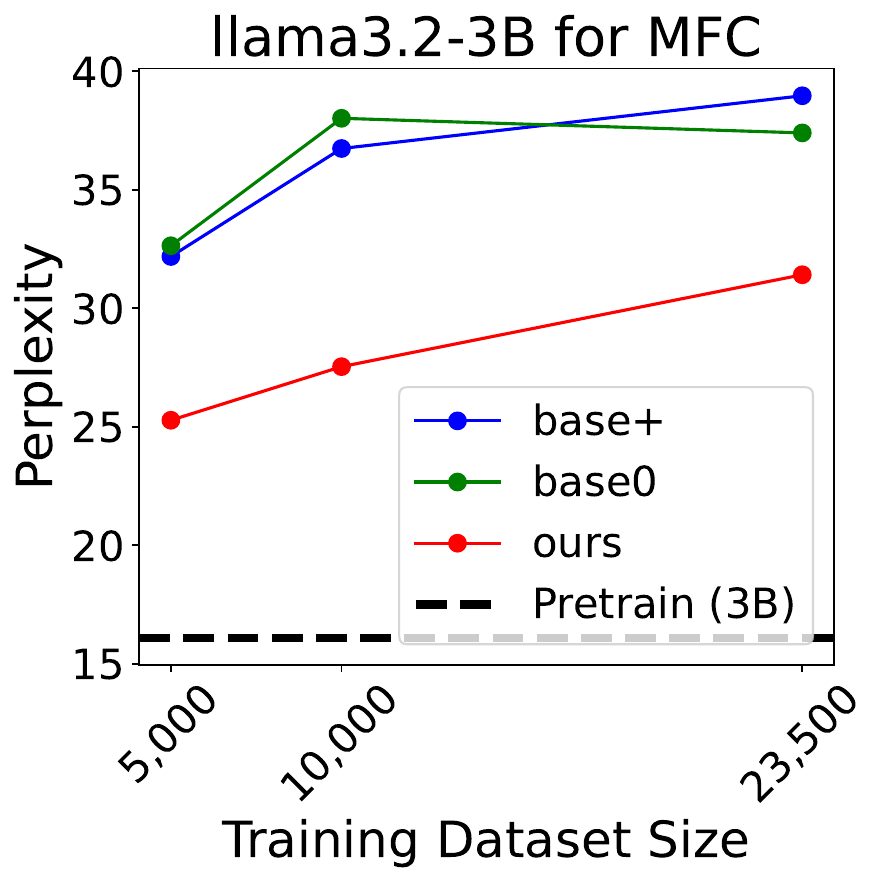}
    \end{minipage}        
    \begin{minipage}[b]{0.240\textwidth}
        \includegraphics[width=\textwidth]{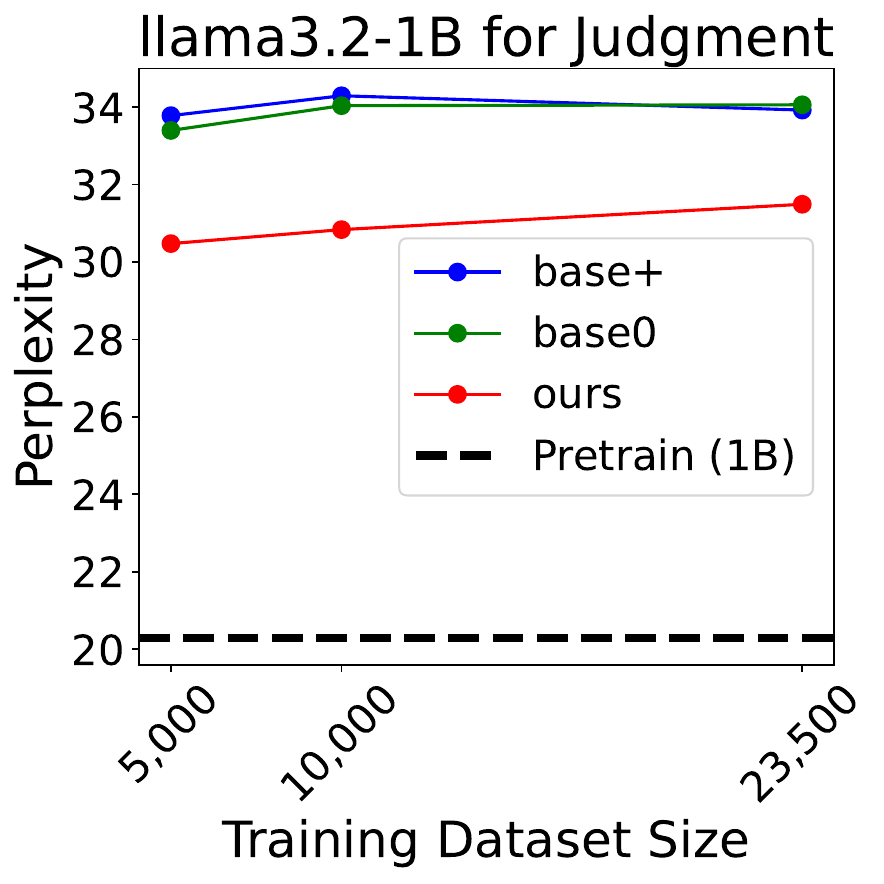}
    \end{minipage}\hfill
    \begin{minipage}[b]{0.240\textwidth}
        \includegraphics[width=\textwidth]{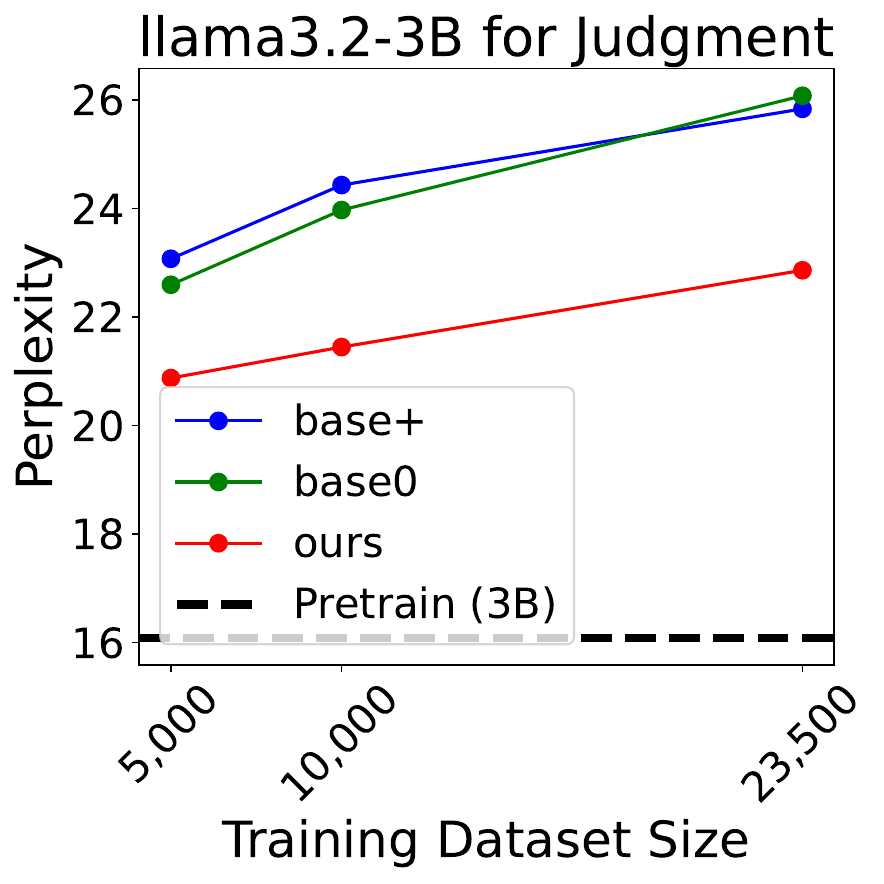}
    \end{minipage}
    \caption{Perplexity of Fine-tuned Models Under Different Moral Reasoning Approaches on the Moral Foundations Classification (MFC) and Moral Judgment Tasks. Our proposed methods result in a much smaller perplexity than baseline methods, suggesting a better language modeling capability. 
    }
    \label{fig:perplexity}
\end{figure}
\paragraph{Perplexity.} Although fine-tuning can introduce forgetting in LLMs~\cite{kirkpatrick2017overcoming,liu2024towards} and degrade their general capabilities,~\citet{liu2025diagnosing} emphasize that fine-tuning for moral reasoning, when not accounting for the pragmatic nature of morality, tends to yield higher perplexity.
This suggests that the fine-tuning discourses are not semantically sufficient for LLMs to learn.
This is also verified in other morality-relevant tasks such as social stereotype mitigation~\cite{liu2025diagnosingperformancetradeoffmoral}.
To further demonstrate that \textbf{our pragmatic inference method closes the pragmatic gap and makes the fine-tuning discourse for moral reasoning more learnable for LLMs}, we calculate the Perplexity of the optimal fine-tuned checkpoints of baseline approaches and our proposed approach.
We use the WikiText dataset~\cite{merity2016pointer} to compute perplexity.
Figure~\ref{fig:perplexity} reports perplexity results across different training dataset sizes and tasks. Models fine-tuned with our pragmatic inference methods consistently exhibit substantially lower perplexity compared to baseline approaches.

\begin{figure}[h!]
    \centering
       
    \begin{minipage}[b]{0.235\textwidth}
        \includegraphics[width=\textwidth]{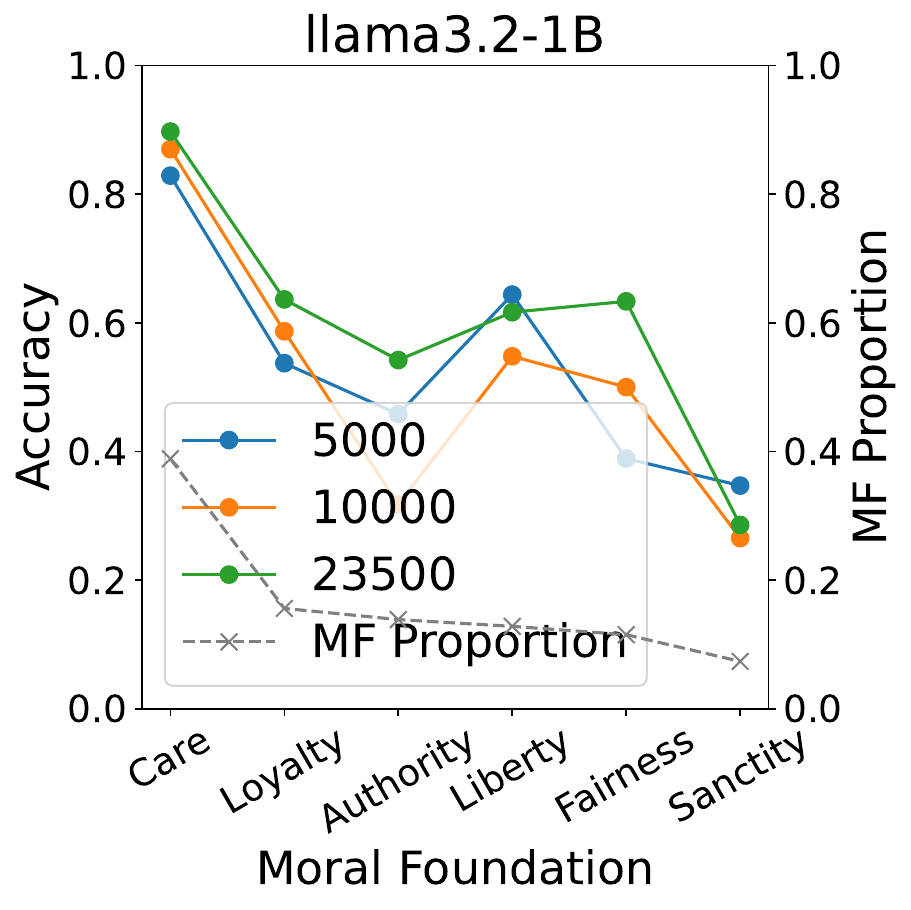}
    \end{minipage}  
    \begin{minipage}[b]{0.235\textwidth}
        \includegraphics[width=\textwidth]{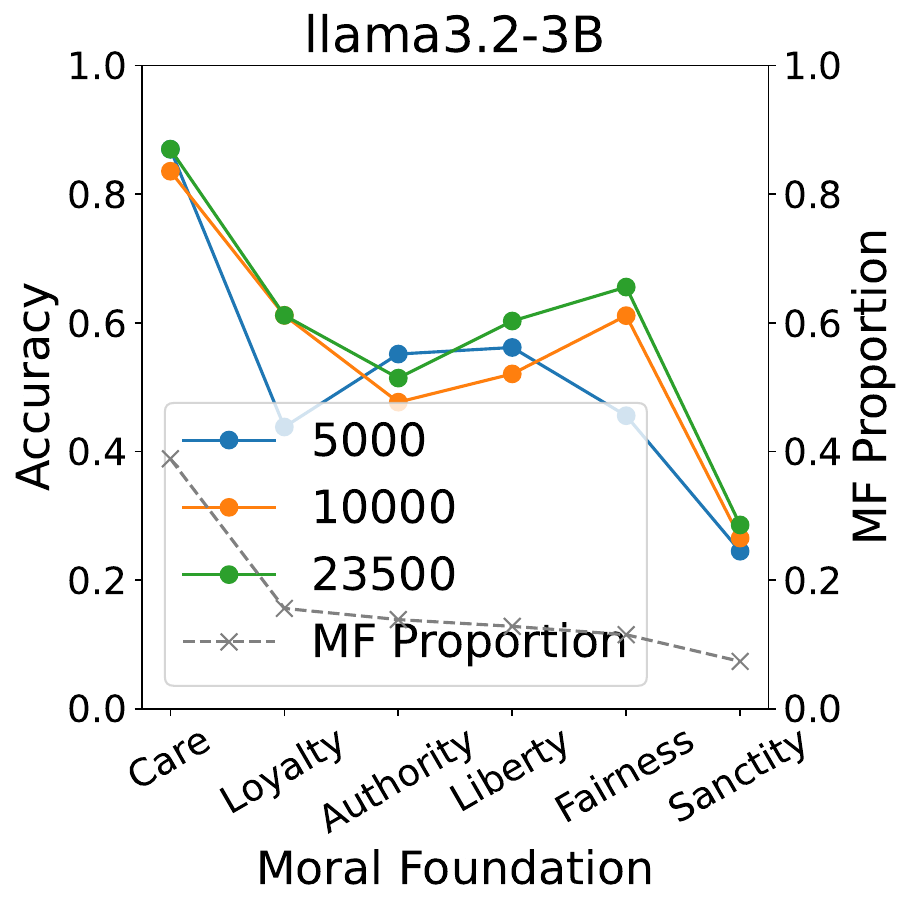}
    \end{minipage}   
    \caption{Performance of Moral Judgment Across Individual MFs. We report results only on test situations involving a single underlying moral foundation for more precise analysis. The black dashed line indicates the proportion of cases corresponding to each moral foundation in the training dataset. The X-axis orders the MFs in ascending order of their proportions.}
    \label{fig:mfwise}
\end{figure}
\paragraph{Moral Foundation-wise Analysis.} 
Since MFs play an essential role in our pragmatic inference approach, we wonder whether and how the different distribution of them may affect the performance of models in the moral judgment task. Figure~\ref{fig:mfwise} illustrates MFs-wise performance, highlighting a correlation between the number of situations of each MF in the training dataset and the eventual accuracy of moral judgment. 
Notably, care accounts for about 40\% of the training data, leading to higher model performance in test situations involving this foundation.
In contrast, performance is much lower for test situations involving the \textit{sanctity} foundation. 
These empirical results suggest a space for our methods to further improve model performance, if given sufficient and balanced annotation data.

\section{Discussion\label{sec:discussion}}
Different from existing approaches that train models to map moral situations to moral judgment, ~\cite{hendrycks2020aligning,forbes2020social,jiang2021can,ziems2022moral,jiang2025investigating}, this paper develops a pragmatic inference approach that bridges the two. The bridge is built to close the pragmatic gap between what is said and what is morally implied. It did so by textualizing the metapragmatic links between the designed components in the inferential process. In other words, the approach leverages the ``internal representations of social variables'' \cite{yang_socially_2025} and facilitates LLMs to learn to draw on them. 
It is in line with previous studies that develop other pragmatic inferential methods for toxic language detection tasks ~\cite{sap2020social,zhou-etal-2023-cobra,chen2025pragmatic}. 

However, it further illuminates the need for metapragmatic links to be incorporated into LLM training. This finding provides another explanation for the effectiveness of step-wise design, such as CoT, in improving LLM performance. In other words, what the step-wise responses present to LLMs may be the meta-links (not necessarily metapragmatic links) that have ``internal representations'' \cite{yang_socially_2025} but need LLMs to learn how to draw on. The finding also echoes previous arguments about the importance of annotating reasoning process, instead of the outcomes only \cite{wolfson_monaco_2026, gao_hierarchical_2022}. 

Moreover, the performance increase that we find across three different tasks and different models suggests a potential generalizability of the current approach in different pragmatic tasks
such as the understanding of metaphors, humor, and sarcasm, which LLMs have thus far been found deficient in \cite{barattieri_di_sanpietro_pragmatic_2023, ruis_goldilocks_2023}. 
For future work, this paper can be extended in two directions: (1) incorporating moral foundation inference, and (2) exploring more practical scenarios involving metaphors and referential expressions in text.

\end{document}